\documentclass{article} 
\usepackage{iclr2024_conference,times}

\usepackage{amsmath,amsfonts,bm}

\def\eqref#1{equation~\ref{#1}}

\def\1{\bm{1}}

\DeclareMathAlphabet{\mathsfit}{\encodingdefault}{\sfdefault}{m}{sl}
\SetMathAlphabet{\mathsfit}{bold}{\encodingdefault}{\sfdefault}{bx}{n}

\usepackage{hyperref}
\usepackage{url}
\usepackage{float}
\usepackage{graphicx}

\title{Mapping Land Naturalness from Sentinel-2 using Deep Contextual and Geographical Priors}

\author{Burak Ekim \& Michael Schmitt \ \\
Department of Aerospace Engineering\\
University of the Bundeswehr Munich\\
Neubiberg, Germany\\
\texttt{\{burak.ekim,michael.schmitt\}@unibw.de} \\
}

\iclrfinalcopy 
\begin{document}

\maketitle

\begin{abstract}
In recent decades, the causes and consequences of climate change have accelerated, affecting our planet on an unprecedented scale. This change is closely tied to the ways in which humans alter their surroundings. As our actions continue to impact natural areas, using satellite images to observe and measure these effects has become crucial for understanding and combating climate change. Aiming to map land naturalness on the continuum of modern human pressure, we have developed a multi-modal supervised deep learning framework that addresses the unique challenges of satellite data and the task at hand. We incorporate contextual and geographical priors, represented by corresponding coordinate information and broader contextual information, including and surrounding the immediate patch to be predicted. Our framework improves the model's predictive performance in mapping land naturalness from Sentinel-2 data, a type of multi-spectral optical satellite imagery. Recognizing that our protective measures are only as effective as our understanding of the ecosystem, quantifying naturalness serves as a crucial step toward enhancing our environmental stewardship.
\end{abstract}

\section{Introduction}
Human beings have always changed the land to suit their needs. From clearing forests to building cities, our actions shape the environment in profound ways. Unfortunately, the cumulative effect of these activities and changes has had serious effects on the planet that we all live in. With ecosystems increasingly impacted by modern human activities, Earth Observation (EO) has emerged as a pivotal tool. Serving as our eyes from above, EO provides a clear picture of our interaction with the natural world, playing a crucial role in both monitoring and mitigating the effects of anthropogenic climate change. Assessing the integrity of natural environments through detailed mapping is crucial for steering restoration initiatives, safeguarding diverse life forms, enhancing ecosystem robustness against human pressures, and upholding the equilibrium vital for the well-being of Earth and its inhabitants. The aspect of modern human influence on the environment includes a range of activities like farming, urban development, industry, and conservation efforts. \cite{sanderson2002} has laid the groundwork for measuring these influences on the environment, taking into account elements including proximity to rivers, settlements, urban areas, and distances from roads, major rivers, or coasts. \cite{kennedy2019managing} also advanced the field by mapping human land alterations globally in 2016, using data on 13 human-induced stressors, uncovering that most of the world's eco-regions show moderate modification. This highlighted the urgent need for conservation efforts in these regions to preserve biodiversity and ecosystem services. \cite{venter2016global} also introduced the first time-consistent Human Footprint maps, integrating various data on human pressures, thus offering insights into the impact of human activity on ecological patterns.

Recently, \cite{naturalness} also introduced the Naturalness Index (NI), a metric ranging from 0 to 100 that measures the natural state of the Earth with a 10-meter resolution. They build upon the foundational work by \cite{sanderson2002}, incorporating land cover penalty factors and accessibility dimension to develop a more refined, high-resolution metric. The NI is defined as a composite measure that includes accessibility through roads, railways, and waterways (how reachable a location is), population density (where people reside), land cover (how we use the land) and the night-time lights seen from space. 

However, traditional approaches sometimes fall short in capturing the complexity of modern human influence. Traditional methods depend on the availability of multiple drivers with low temporal frequency and vary in spatial detail, impeding a comprehensive understanding of environmental impacts. Further, they fuse the drivers based on rule-based methods with heuristics without considering geographic (the geolocation of the landscape) and contextual (the area surrounding the landscape) factors, underscoring the need for approaches that account for human influence by considering spatial correlation and the effects of nearby activities. Lastly, while the drivers used to assess this impact (e.g., OpenStreetMap \cite{OpenStreetMap}) are widely available, their completeness varies across regions, typically being more comprehensive in data-rich regions \cite{osm2}, leading to biased assessments of environmental impact, potentially overlooking or underestimating the effects in less documented areas. 

Our research fills this gap by proposing a multimodal supervised deep learning approach mapping land naturalness at the pixel level using a single Sentinel-2 image, eliminating the need for multiple drivers and their heuristic combination rules. Our method incorporates geographic coordinates and contextual information surrounding the patch of interest, tailored to meet the nuanced demands of mapping land naturalness using satellite images.

\section{Modelling Spatial Correlation}

CNNs excel in spatial data processing through their structure, yet struggle to fully capture the intricate spatial correlations and distant dependencies crucial for satellite imagery analysis, often overlooking vital broader context \cite{LeCun2015DeepLearning, Krizhevsky2012ImageNetCW}. Such limitations are starkly evident in satellite imagery, where understanding the complex web of spatial relationships and dependencies is essential for accurate interpretation across diverse landscapes \cite{tuia2023}. In EO, a profound understanding of environmental phenomena requires analyzing not just direct observations but also the broader spatial contexts that influence them, such as in tasks like PM2.5 estimation, vegetation health monitoring, urban heat island assessment, and coastal erosion analysis. The complexity of these tasks underscore the need for models with an inductive bias toward spatial autocorrelation, adhering to Tobler’s law, "Everything is related to everything else, but nearby things are more related than distant things," to accurately depict the nuanced spatial relationships in satellite imagery. Our method enhances CNNs by integrating this inductive bias and additional context and geographic data, filling gaps left by traditional models. The method captures both the immediate area of interest and the intricate spatial dynamics of the surrounding landscape, providing a more holistic view that's crucial for the distinct modality of satellite imagery.

\section{Methodology}
\begin{figure}[htbp]
\begin{center}
\includegraphics[scale=0.5]{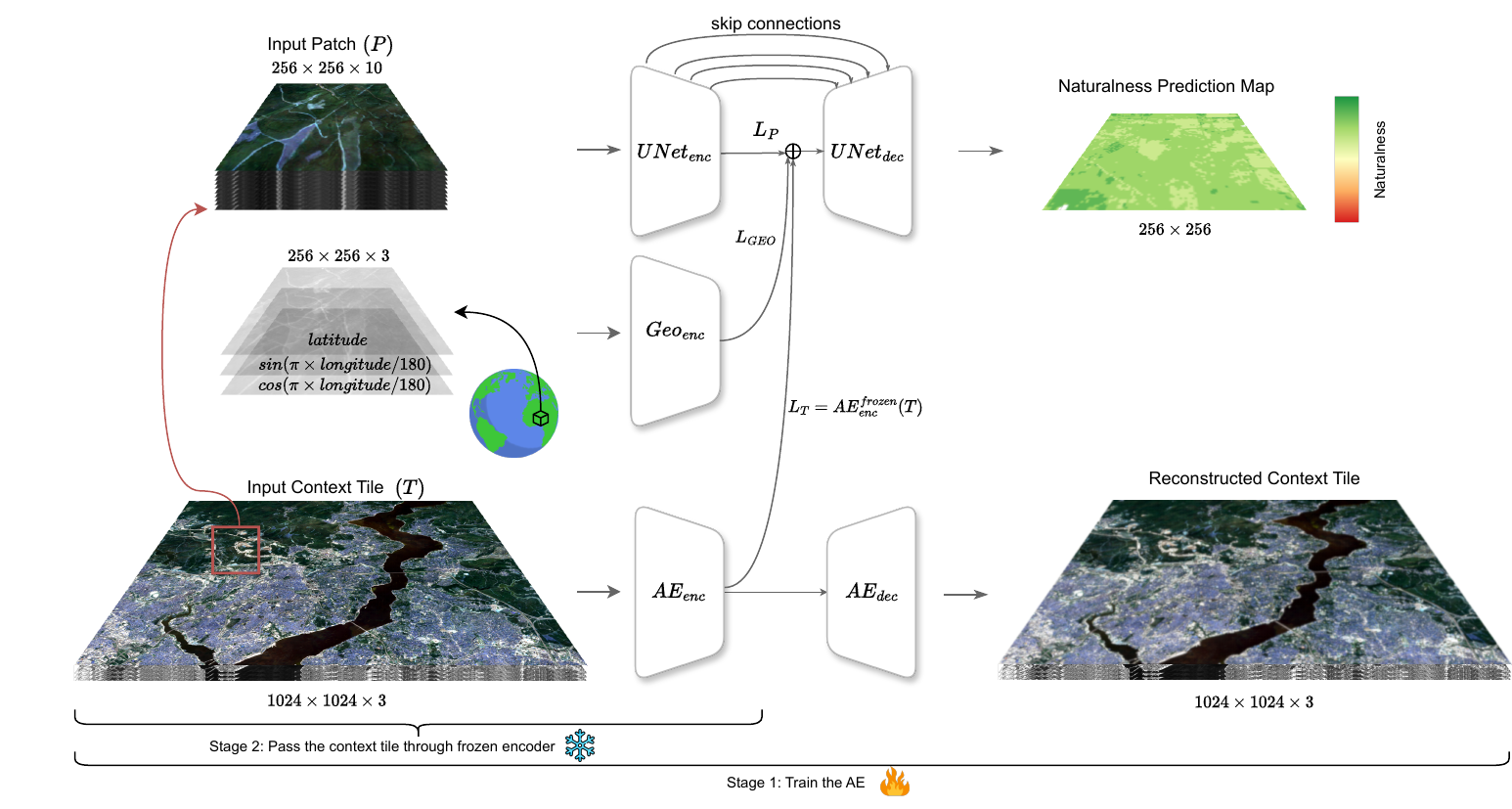}
\end{center}
\caption{The proposed framework works as follows: Initially, an Autoencoder (AE) is trained to reconstruct the input context tiles, encoding the contextual information into a latent space. The broader context tiles pass through the now frozen AE encoder \(AE_{enc}\), while the smaller patches cropped from these tiles are fed to the UNet encoder \(UNet_{enc}\) and the encoded coordinates are fed to the \(Geo_{enc}\). The three high-dimensional latent representations obtained from all three encoders are then channel-wise concatenated and input into \(UNet_{dec}\) to produce the naturalness prediction map.}

\label{fig:flow}
\end{figure}

In this framework, an image is defined as \( I \in \mathbb{R}^{H \times W \times C} \), where \( H \), \( W \), and \( C \) denote the height, width, and number of spectral channels, respectively. For a given small patch \( P \) with dimensions \( h \times w \), extracted from \( I \), the goal is to predict a naturalness map \( M \in \mathbb{R}^{h \times w} \) where each pixel in \( M \) corresponds to a naturalness score. The high-level overview of the framework is illustrated in Figure \ref{fig:flow}. In the following, we break down the features of the framework:

\textbf{Pixel-wise regression model}: For pixel-wise regression, the vanilla UNet \cite{ronneberger2015unet} architecture is used, which consists of an encoder $\text{UNet}_{enc}$ and a decoder $\text{UNet}_{dec}$. The encoder, $\text{UNet}_{enc}$, processes the patch $P$ and applies convolutional operations, followed by batch normalization, ReLU, and max pooling, to create an embedding $L_{P} = \text{UNet}_{enc}(P)$. The decoder, $\text{UNet}_{dec}$, uses transposed convolutional layers and mirrors the encoder. Furthe, the decoder generates the predicted naturalness map $M$ through upsampling, where $M = \text{UNet}_{dec}(L_{P})$. The UNet outputs raw logits as the prediction output.

\textbf{Cyclic encoding of coordinates}: Several studies have previously employed encoding techniques for geographic coordinates \cite{geographyaware, SatCLIP, Lang2023}. The reason for cyclic encoding of longitude coordinates stems from the fact that the Earth's longitude values are continuous and wrap around at +180 and -180 degrees, forming a loop. However, traditional scalar representations of longitude fail to capture this continuity; for example, +179 and -179 degrees are geographically close but appear numerically distant. By transforming longitude into a cyclic format using $\cos(2 \pi \times \text{longitude}/180)$ and $\sin(2 \pi \times \text{longitude}/180)$, we effectively map the continuous nature of these coordinates onto a circle, where positions at +180 and -180 degrees are represented as the same point. Further, the use of both sine and cosine together is motivated by the fact that there is a $\pi/2$ phase difference, which prevents loss of information at zero-crossings.

\textbf{Image-to-Image translation model}: Incorporating additional information such as scale or context has been extensively explored, with Autoencoders (AEs) being a notable example. The AE in this study is designed for image reconstruction on larger context tiles $T$. It encodes broader spatial information into a latent space, $L_{T} = \text{AE}{enc}(T)$, and reconstructs the high-dimensional embedding with its decoder, $\text{AE}{dec}$. The AE's encoder segment has convolutional layers with increasing channel sizes, similar to the UNet encoder, followed by pooling layers to reduce spatial dimensions.

\textbf{Bringing it all together}: The Autoencoder's encoder, $\text{AE}{enc}$, processes the larger context tile $T \in \mathbb{R}^{H \times W \times C}$ to produce a latent representation $L{T}$, encapsulating the broader spatial context ($L_{T} = \text{AE}{frozenenc}(T)$). Concurrently, the UNet encoder maps $P$ into a latent space representation ($L{P} = \text{UNet}{enc}(P)$). A dedicated encoder, $\text{Geo}{enc}$, maps the coordinates into a high-dimensional feature vector ($L_{C} = \text{Geo}{enc}(C)$). The latent representations are then concatenated channel-wise to create a combined feature map ($L{combined} = L_{P} \oplus L_{C} \oplus L_{T}$). This feature map is then passed through the UNet decoder to form the naturalness map ($M = \text{UNet}{dec}(L{combined})$). In summary, we integrate auxiliary supervision on an embedding level, introducing broader contextual information from the larger context tile and spatial-geographic relation from the encoded coordinates. The latent space learned by the autoencoder can capture high-level abstractions and relationships in the data, which can be useful for understanding spatial relationships from the compressed representations of raw data. Further, encoding the coordinates ensures that our model accurately interprets the geographic continuity and layout of the Earth's surface, uncovering spatial relationships and patterns. The motivation behind these methods is the understanding that the naturalness state of a pixel is significantly influenced by its geographic location on Earth and the characteristics of its surrounding environment.
\section{Experimental Setup and Results}
We form the dataset used in this study by coupling the 10-band Sentinel-2 patches of MapInWild \cite{grsm_wild} with newly created corresponding NI patches, as shown in Figure \ref{fig:ni_sample}. MapInWild's diverse and global coverage of both anthropogenic and wild landscapes offer an ideal foundation for integrating the naturalness annotations, enabling a harmonious and complementary relationship.  

\begin{figure}
    \centering
    \includegraphics[width=1\linewidth]{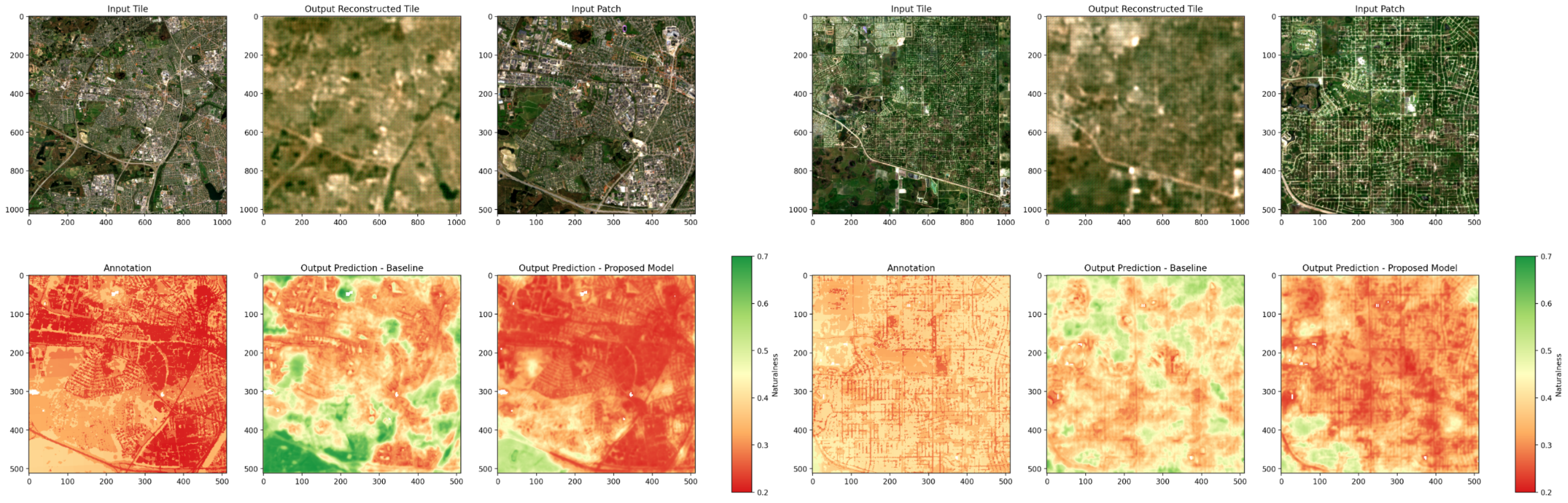}
    \caption{Inference results of the patches with dataset IDs 900000061 and 900000068. The areas cover parts of Cape Coral, Florida, USA (left), and Copenhagen, Denmark (right). For each prediction, the first row shows the context tile ($T$), its reconstruction, and the main input patch ($P$). The second row shows the naturalness annotation and the prediction results of the baseline and proposed models.}
    \label{fig:pred_87}
\end{figure}

Two experimental setups are formed to compare the baseline model (a plain UNet) with the proposed model (as illustrated in Figure \ref{fig:flow}). Quantitative results are tabulated in Table \ref{tab:ablation_study}, and qualitative  results are provided in Figure \ref{fig:pred_87}. The model produces a prediction map for a given patch $P$, its geographic coordinates, and the context $T$ images. The patches are cropped from the center of context images. The reconstructed tiles are presented to demonstrate the performance of the AE.  Details on the dataset and implementation details can be found in Appendix A.

\begin{table}[h]
\centering
\begin{tabular}{lccc}
\hline
\textbf{Features} & \textbf{MAE $\downarrow$} & \textbf{MSE $\downarrow$} & \textbf{MSSIM $\uparrow$} \\ \hline
Baseline & 0.24 & 0.29 & 0.78 \\
+ Coordinate and Context & 0.14 & 0.25 & 0.89 \\ \hline
\end{tabular}
\caption{The comparison study between the baseline and the proposed model.}
\label{tab:ablation_study}
\end{table}

This comparison suggests that the proposed model enhances predictive performance, particularly for urban structures. The inclusion of geographic coordinates and broader context information seems to benefit the main regression task, with the auxiliary embeddings providing additional insights drawn from spatial relationships and broader contextual information. The evaluation metrics reflect our model's capability to gauge the degree of naturalness in satellite imagery; more accurate predictions translate to a more precise understanding of the landscape's state, crucial for improving our conservation strategies.

\section{Discussion and Conclusion}
Earth observation has fundamentally changed the way we address climate change, enabling us to map the integrity of natural landscapes, a concern increasingly under threat due to anthropogenic pressures. Mapping the integrity of natural landscapes is key to driving conservation, restoring damaged ecosystems, preserving biodiversity, and bolstering resilience to human impacts, ensuring the balance critical for the health of our planet and our survival. 

We propose a framework predicting the naturalness state of a landscape from multi-spectral optical satellite imagery using the naturalness index as an annotation source. The naturalness index quantifies the disturbance on a landscape by heuristically fusing pressures that account for modern human influence. With the proposed framework, we bypass traditional heuristic methods for data fusion and eliminate the dependency on the availability of specific drivers. This enables dense prediction at a consistent spatial and potentially high temporal resolution, requiring only a single satellite image.

The informed inclusion of coordinate, local, and global contextual information is specifically designed to meet the task's inherent demands, which heavily rely on geographic and contextual nuances for accurate naturalness quantification. We hope our work offers an insight into the mapping of the modern human footprint, providing a new perspective for crafting more effective and informed conservation strategies.

\bibliography{iclr2024_conference}

\begin{thebibliography}{15}
\providecommand{\natexlab}[1]{#1}
\providecommand{\url}[1]{\texttt{#1}}
\expandafter\ifx\csname urlstyle\endcsname\relax
  \providecommand{\doi}[1]{doi: #1}\else
  \providecommand{\doi}{doi: \begingroup \urlstyle{rm}\Url}\fi

\bibitem[Ayush et~al.(2022)Ayush, Uzkent, Meng, Tanmay, Burke, Lobell, and Ermon]{geographyaware}
Kumar Ayush, Burak Uzkent, Chenlin Meng, Kumar Tanmay, Marshall Burke, David Lobell, and Stefano Ermon.
\newblock {Geography-Aware Self-Supervised Learning}.
\newblock \emph{arXiv:2011.09980}, 2022.

\bibitem[Barrington-Leigh \& Millard-Ball(2017)Barrington-Leigh and Millard-Ball]{osm2}
Christopher Barrington-Leigh and Adam Millard-Ball.
\newblock The world’s user-generated road map is more than 80
\newblock \emph{PLOS ONE}, 12\penalty0 (8):\penalty0 1--20, 08 2017.
\newblock \doi{10.1371/journal.pone.0180698}.

\bibitem[Ekim et~al.(2021)Ekim, Dong, Rashkovetsky, and Schmitt]{naturalness}
Burak Ekim, Zeyu Dong, Dmitry Rashkovetsky, and Michael Schmitt.
\newblock The naturalness index for the identification of natural areas on regional scale.
\newblock \emph{International Journal of Applied Earth Observation and Geoinformation}, 105:\penalty0 102622, 2021.
\newblock ISSN 1569-8432.

\bibitem[Ekim et~al.(2023)Ekim, Stomberg, Roscher, and Schmitt]{grsm_wild}
Burak Ekim, Timo~T. Stomberg, Ribana Roscher, and Michael Schmitt.
\newblock {MapInWild: A remote sensing dataset to address the question of what makes nature wild [Software and Data Sets]}.
\newblock \emph{IEEE Geoscience and Remote Sensing Magazine}, 11\penalty0 (1):\penalty0 103--114, 2023.

\bibitem[Kennedy et~al.(2019)Kennedy, Oakleaf, Theobald, Baruch-Mordo, and Kiesecker]{kennedy2019managing}
Christina~M Kennedy, James~R Oakleaf, David~M Theobald, Sharon Baruch-Mordo, and Joseph Kiesecker.
\newblock Managing the middle: A shift in conservation priorities based on the global human modification gradient.
\newblock \emph{Global Change Biology}, 25\penalty0 (3):\penalty0 811--826, 2019.

\bibitem[Klemmer et~al.(2023)Klemmer, Rolf, Robinson, Mackey, and Rußwurm]{SatCLIP}
Konstantin Klemmer, Esther Rolf, Caleb Robinson, Lester Mackey, and Marc Rußwurm.
\newblock {SatCLIP: Global, General-Purpose Location Embeddings with Satellite Imagery}.
\newblock \emph{arXiv:2311.17179}, 2023.

\bibitem[Krizhevsky et~al.(2017)Krizhevsky, Sutskever, and Hinton]{Krizhevsky2012ImageNetCW}
Alex Krizhevsky, Ilya Sutskever, and Geoffrey~E. Hinton.
\newblock {ImageNet classification with deep convolutional neural networks}.
\newblock \emph{Commun. ACM}, 60\penalty0 (6):\penalty0 84–90, may 2017.
\newblock ISSN 0001-0782.

\bibitem[Lang et~al.(2023)Lang, Jetz, Schindler, and Wegner]{Lang2023}
Nico Lang, Walter Jetz, Konrad Schindler, and Jan~Dirk Wegner.
\newblock A high-resolution canopy height model of the earth.
\newblock \emph{Nature Ecology \&; Evolution}, 7\penalty0 (11):\penalty0 1778–1789, September 2023.
\newblock ISSN 2397-334X.

\bibitem[LeCun et~al.(2015)LeCun, Bengio, and Hinton]{LeCun2015DeepLearning}
Yann LeCun, Yoshua Bengio, and Geoffrey Hinton.
\newblock Deep learning.
\newblock \emph{Nature}, 521\penalty0 (7553):\penalty0 436--444, 2015.

\bibitem[{OpenStreetMap Contributors}(2023)]{OpenStreetMap}
{OpenStreetMap Contributors}.
\newblock {OpenStreetMap}.
\newblock \url{https://www.openstreetmap.org}, 2023.
\newblock Accessed: [20.12.2023].

\bibitem[Ronneberger et~al.(2015)Ronneberger, Fischer, and Brox]{ronneberger2015unet}
Olaf Ronneberger, Philipp Fischer, and Thomas Brox.
\newblock {U-Net: Convolutional Networks for Biomedical Image Segmentation}.
\newblock \emph{arXiv:1505.04597}, 2015.

\bibitem[Sanderson et~al.(2002)Sanderson, Jaiteh, Levy, Redford, Wannebo, and Woolmer]{sanderson2002}
Eric~W Sanderson, Malanding Jaiteh, Marc~A Levy, Kent~H Redford, Antoinette~V Wannebo, and Gillian Woolmer.
\newblock The human footprint and the last of the wild: the human footprint is a global map of human influence on the land surface, which suggests that human beings are stewards of nature, whether we like it or not.
\newblock \emph{BioScience}, 52\penalty0 (10):\penalty0 891--904, 2002.

\bibitem[Tuia et~al.(2023)Tuia, Schindler, Demir, Camps-Valls, Zhu, Kochupillai, Džeroski, van Rijn, Hoos, Frate, Datcu, Quiané-Ruiz, Markl, Saux, and Schneider]{tuia2023}
Devis Tuia, Konrad Schindler, Begüm Demir, Gustau Camps-Valls, Xiao~Xiang Zhu, Mrinalini Kochupillai, Sašo Džeroski, Jan~N. van Rijn, Holger~H. Hoos, Fabio~Del Frate, Mihai Datcu, Jorge-Arnulfo Quiané-Ruiz, Volker Markl, Bertrand~Le Saux, and Rochelle Schneider.
\newblock Artificial intelligence to advance earth observation: a perspective.
\newblock \emph{arXiv:1506.01186}, 2023.

\bibitem[{UNEP-WCMC} \& {IUCN}(2023){UNEP-WCMC} and {IUCN}]{WDPA}
{UNEP-WCMC} and {IUCN}.
\newblock {Protected Planet: The World Database on Protected Areas (WDPA)}, 2023.
\newblock URL \url{https://www.protectedplanet.net/en}.
\newblock Accessed: [20.12.2023].

\bibitem[Venter et~al.(2016)Venter, Sanderson, Magrach, Allan, Beher, Jones, Possingham, Laurance, Wood, Fekete, et~al.]{venter2016global}
Oscar Venter, Eric~W Sanderson, Ainhoa Magrach, James~R Allan, Jutta Beher, Kendall~R Jones, Hugh~P Possingham, William~F Laurance, Peter Wood, Bal{\'a}zs~M Fekete, et~al.
\newblock Global terrestrial human footprint maps for 1993 and 2009.
\newblock \emph{Scientific data}, 3\penalty0 (1):\penalty0 1--10, 2016.

\end{thebibliography}
\bibliographystyle{iclr2024_conference}


\newpage
\appendix
\section{Appendix}
\subsection{Dataset}
We extend the MapInWild dataset \cite{grsm_wild}, which is designed for the purpose of mapping wilderness areas using satellite images and auxiliary geospatial data. The dataset comprises 1018 regions sampled from global landmass, with each region formatted as 1920 $\times$ 1920 pixels. MapInWild incorporates a range of sensors, including dual-polarization imagery from Sentinel-1, 10-channel multi-spectral imagery from Sentinel-2, night-time light data from the Visible Infrared Imaging Radiometer Suite, and the ESA WorldCover map. Each MapInWild sample is annotated with polygons from the World Database on Protected Areas \cite{WDPA}, providing extensive information on global conservation regions. To achieve a balanced and representative depiction of the Earth's natural environments, MapInWild integrates a semi-automated process guided by climate zone and land cover proxies to promote representability. Additionally, the dataset encompasses a manually collected set of locations with varying degrees of modern human activity, ensuring a proper utilization of naturalness continuum.

\begin{figure}[H]
\begin{center}
\includegraphics[scale=0.27]{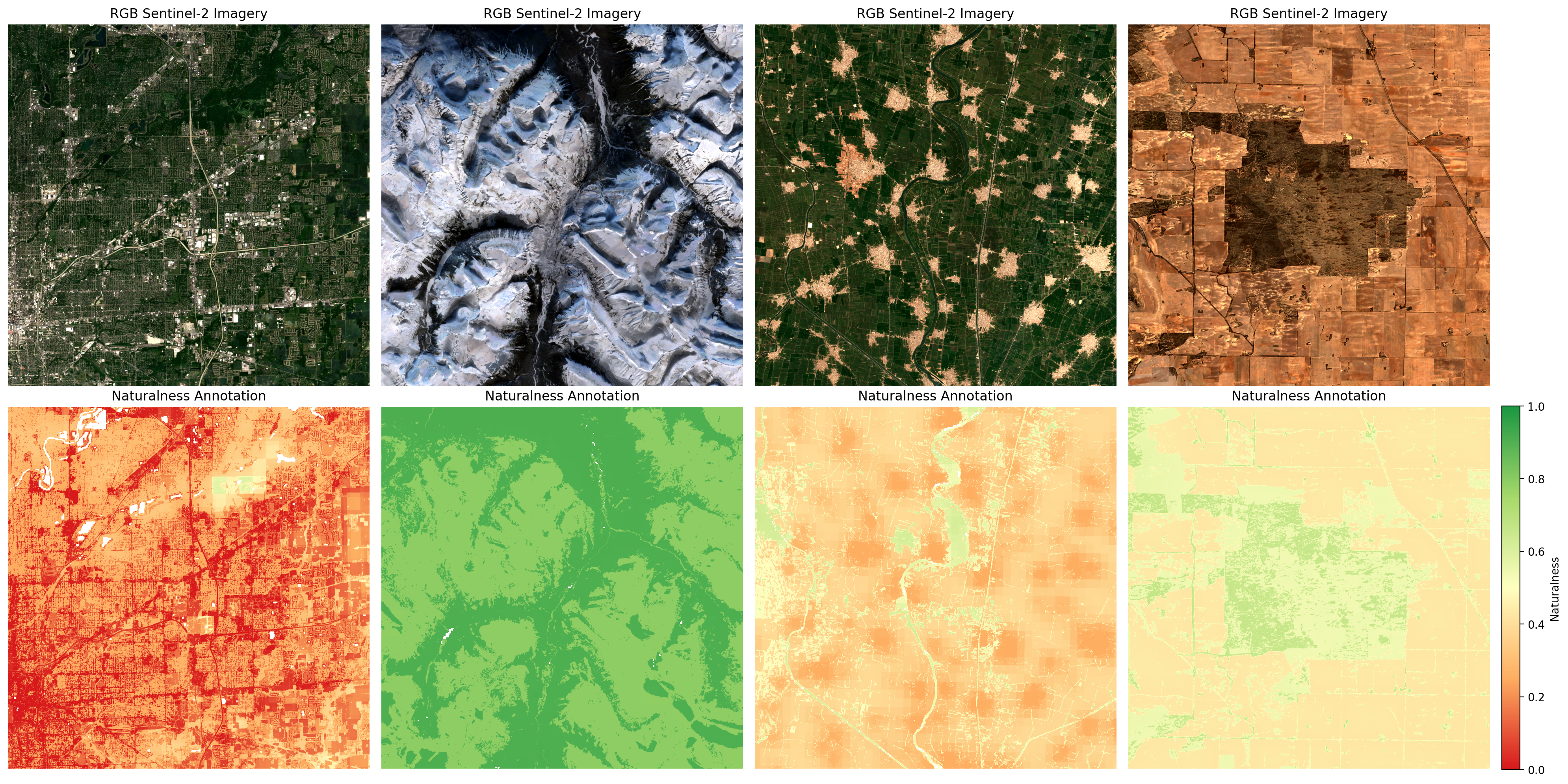}
\end{center}
\caption{Sample Sentinel-2 imageries in true color and their corresponding Naturalness Index maps. Their IDs are 900000093, 4193, 900000027, and 314770.}
\label{fig:ni_sample}
\end{figure}

\subsection{Implementation Details}
The UNet and AE architectures both consist of consecutive blocks with channel sizes of 64, 128, 256, and 512. Broader context tiles are shaped as 1024 $\times$ 1024 $\times$ 3 (B4, B4, B2) and the UNet processes patches of 256 $\times$ 256 $\times$ 10. The cyclically encoded geographic coordinates, shaped as 256 $\times$ 256 $\times$ 3, are fed into an encoder mimicking UNet's. We use Adam optimizer configured with a learning rate of $10^{-4}$ and a weight decay of $10^{-3}$. A cosine annealing learning rate with warm restarts is applied, with initial parameters $T_0$ set to 10 and $T_{mult}$ set to 2. Model training is conducted with a batch size of 16, using 16-bit precision for computation, and includes a weighted sampler to tackle data imbalance. The training set is augmented with vertical and horizontal transformations, and random erasing of image parts. We employ early stopping, with a patience parameter of 15 epochs, based on validation loss. Mean Absolute Error loss is used for calculations, excluding water bodies to refine focus on terrestrial areas. Two experimental setups are formed to compare the baseline model with the proposed model (as illustrated in Figure \ref{fig:flow}). Qualitative results are tabulated in Table \ref{tab:ablation_study}, and quantitative results are provided in Figure \ref{fig:pred_87}. 

\end{document}